\documentclass{elsarticle}

\usepackage{amsmath}
\usepackage{graphicx}
\usepackage{csquotes}
\usepackage{amsfonts}
\usepackage{multicol}
\usepackage{tabularx}
\usepackage{appendix}
\begin{document}
\begin{frontmatter}

\title{Fast shared response model for fMRI data}
\author[1]{Hugo Richard\corref{correspondingauthor}}
\cortext[correspondingauthor]{Corresponding author}
\ead{hugo.richard@inria.fr}
\author[1]{Lucas Martin}
\author[1]{Ana Lu\'{i}sa Pinho}
\author[2]{Jonathan Pillow}
\author[1]{Bertrand Thirion}
\address[1]{Parietal team, Inria, Neurospin, CEA, Universit\'{e} Paris-Saclay, France}
\address[2]{Princeton Neuroscience Institute, Princeton University, USA}

\begin{abstract}
The shared response model provides a simple but effective framework to analyse fMRI data of subjects 
exposed to naturalistic stimuli.
However when the number of subjects or runs is large, fitting the model requires a large amount of memory and computational power, 
which limits its use in practice. 
In this work, we introduce the FastSRM algorithm that relies on an intermediate atlas-based representation. 
It provides considerable speed-up in time and memory usage, hence it allows easy and fast large-scale
analysis of naturalistic-stimulus fMRI data.
Using four different datasets, we show  that our method matches the performance
of the original SRM algorithm while 
being about 5x faster and 20x to 40x more memory efficient.
Based on this contribution, we use FastSRM to predict age from movie watching data on the CamCAN 
sample.
Besides delivering accurate predictions (mean absolute error of 7.5 years), FastSRM extracts topographic
patterns that are predictive of age, demonstrating that brain activity during free perception reflects age.
\end{abstract}

\end{frontmatter}

\section{Introduction}
When exposed to naturalistic stimuli (e.g. movie watching), subjects'
experience is closer to their every-day life than with classical
psychological experiments.
This makes naturalistic paradigms an attractive class of
stimulation protocols for brain imaging.
While there is a broad interest in understanding how the brain reacts
in such ecological conditions, the recorded brain activity is
difficult to analyse.
Standards methods such as the general linear model \cite{poline2012general} require the experimenter to construct a design matrix that models features of the presented stimuli across time. 
Such design matrices are notoriously difficult to construct for naturalistic stimuli as one has to rely on manual annotations (see \cite{huth2012continuous}) or deep learning techniques (see e.g. \cite{gucclu2017increasingly}, \cite{eickenberg2017seeing}, \cite{richard2018optimizing} or \cite{gucclu2015deep}) that are hard to use, and provide high-dimensional, cumbersome models of the stimulus.

\cite{hasson2004intersubject} has shown that brains exposed to the same natural stimuli exhibit synchronous activity.
The shared response model (SRM) \cite{chen2015reduced} models this behaviour and extracts a common response from different subjects exposed to the same stimuli and subject-specific spatial components.
The resulting shared response can serve as a design matrix, while the spatial components are naturally seen as weighting factors.

SRM has initially been designed to work within regions of interest using few subjects.
It has been used in \cite{chen2015reduced} to transfer knowledge between subjects allowing precise location of a 15s time segment of fMRI data from a left-out subject using data from other subjects exposed to the same stimuli. 

A first big step towards more scalability has been made in \cite{anderson2016enabling} reducing fitting time and memory requirements by several orders of magnitude thanks to a smart use of the inversion lemma. 
After this improvement, studies using larger regions of interest have emerged such as \cite{vodrahalli2018mapping}, which uses SRM to predict text embeddings from fMRI data or \cite{chen2017shared}, which shows that shared memories also come with shared structure in neural activity.
However, when using full brain data and a large number of subjects and runs, computational costs are still very high. Memory requirements are difficult to meet since all data have to be loaded in memory and since all full brain spatial components of all subjects are updated at each iteration, this leads to a heavy computational burden.

Fortunately, these high costs can be reduced. Intuitively, since the shared response lives in a reduced space, a compressed representation of the input is good-enough to find a suitable estimate.
With the use of off-the-shelf atlases and careful memory management we implement this idea and build FastSRM.

FastSRM is scalable with the number of runs and subjects. Fitting time and memory requirements are reduced considerably, making it possible to fit FastSRM using a laptop in a reasonable amount of time.
%
FastSRM makes large-scale analysis of movie-watching fMRI fast and easy. We demonstrate its usefulness on CamCAN data where we predict age from movie watching data with a mean absolute error (MAE) of 7.5 years.
We also show in an encoding experiment that FastSRM's ability to transfer data between subjects is superior to current implementations (\cite{anderson2016enabling}).  

Our code is freely available at \url{https://github.com/hugorichard/brainiak/tree/fastsrm}.

There is a long history of using latent factor models for fMRI data analysis.
ICA was first applied on fMRI data in \cite{mckeown1998analysis} as an alternative to the generalized linear (GLM) model.
A few years later, \cite{beckmann2004probabilistic} proposed Probabilistic ICA reducing overfitting problems of original ICA by introducing a Gaussian noise model and low rank structure.
In order to be able to compare patterns across subjects, ICA can be applied on time-wise concatenated data \cite{calhoun2001method}. 
In 2010, \cite{varoquaux2010group} introduces CanICA that yields more stable group decompositions than previous group ICA methods (\cite{calhoun2001method}, \cite{beckmann2005tensorial} and \cite{smith2004advances}).

Since then, a number of different factor models have been applied to fMRI data such as non-negative matrix factorization \cite{xie2017decoding} or dictionary learning \cite{varoquaux:inria-00588898}.
All models enforce constrains on the data that are more or less realistic. \cite{abraham} shows that total variation constrains on spatial components yield accurate and stable decompositions across subjects. However, with high quantities of data, the impact of such regularizations on the result vanishes (see \cite{dohmatob2016learning}) and in this case one should therefore favor efficient algorithms (such as the online dictionary learning implementation of \cite{mensch2016dictionary}).
In practice, most factor algorithms have been used to derive atlases from rest data. But other applications exist. In \cite{varoquaux2013cohort}, dictionary learning is used to derive an atlas from task data (contrast maps) and ICA is used in \cite{calhoun2004alcohol} and \cite{calhoun2002different} to study respectively the effect of alcohol and speed on driving behavior thanks to a simulated driving protocol.

SRM has first been introduced in \cite{chen2015reduced}. Like \cite{lashkari2009exploratory}, it learns subject-specific atlases and extracts a shared functional representation from the fMRI data of multiple subjects exposed to the same stimuli.
It relies on the hypothesis that orthonormality constrains put on spatial components allow for the extraction of more interpretable spatial components from naturalistic-stimulus fMRI data. %
SRM is not sensitive to the spatial variability of activated area between subjects since the only common features across participants are time courses making up the shared representation.
It is thus related to template estimation methods such as hyperalignment \cite{guntupalli2016model} or optimal transport \cite{DBLP:conf/ipmi/BazeilleRJT19}.
Many variants of SRM exists. \cite{chen2016convolutional} uses a convolutional autoencoder to preserve spatial locality, \cite{shvartsman2017matrix} uses matrix normal prior on spatial components and shared response,  \cite{turek2018capturing} adds a subject-specific component and \cite{turek2017semi} gives a semi-supervised version of SRM.
All of these methods suffer from the efficiency issues already mentioned for standard SRM.
Note that, unless specified otherwise, whenever we refer to SRM, we refer to the current most efficient implementation of ProbSRM as provided by \cite{anderson2016enabling} which is freely available in the brainiak library \url{https://github.com/brainiak/brainiak}.

FastSRM needs atlases with a sufficient number of regions to work. Good atlases of various size are available in the literature such as Basc (up to 444 parcels) \cite{bellec2010multi}, Shaeffer (up to 800 parcels) \cite{schaefer2017local} or MODL (up to 1024 parcels) \cite{mensch2018extracting}.
We show that taking any such atlas yields quantitatively similar results.

\section{Material and methods}
\subsection{The shared response model (SRM)}
The shared response model is a latent factor model. The brain images of subject $i$ during run $s$ are stored in a matrix $\mathbf{X}^{(s)}_i \in\mathbb{R}^{t \times v}$ where $v$ is the number of voxels and $t$ the number of acquired brain images.
At time $\tau$, the brain volume $\mathbf{X}^{(s)}_i[\tau]$  is modeled as a weighted sum of $k$ orthonormal spatial components stored in $\mathbf{W}_i \in\mathbb{R}^{k \times v}$.

Keeping things simple, we assume that all $n$ subjects have the same
number $v$ of voxels and all $m$ runs have the same number $t$ of
timeframes.
The extension to the more general case where each run has its own
number of timeframes and each subject its own number of voxels is
straightforward. In our implementation runs can have different number
of timeframes but each subject has the same number of voxels.
Typical values for the number of voxels $v$, the number of timeframes
$t$, the number of runs $m$, the number of subjects $n$ and the
number of components $k$ are given in Table~\ref{tab:typical}.
Let us introduce the following notations:
\begin{itemize}

\item For each subject $i$, the concatenation $\mathbf{X}_i \in \mathbb{R}^{mt \times v}$ of the acquisition data for all runs:
\begin{equation*}
\mathbf{X}_i = 
\begin{bmatrix}
	\mathbf{X}^{(1)}_i \\
	\mathbf{X}^{(2)}_i \\
	\vdots \\
	\mathbf{X}^{(m)}_i
\end{bmatrix}
\end{equation*}

\item The concatenation $\mathbf{X} \in \mathbb{R}^{mt \times nv}$ of the brain acquisition of all subjects for all runs:
\begin{equation*}
\mathbf{X} = 
\begin{bmatrix}
    \mathbf{X}^{(1)}_1 & \mathbf{X}^{(1)}_2 & \cdots & \mathbf{X}^{(1)}_n \\
    \mathbf{X}^{(2)}_1 & \mathbf{X}^{(2)}_2 & \cdots & \mathbf{X}^{(2)}_n\\
    \vdots & \vdots &\ddots & \vdots\\
    \mathbf{X}^{(m)}_1 & \mathbf{X}^{(m)}_2 & \cdots & \mathbf{X}^{(m)}_n \\
\end{bmatrix}
\end{equation*}

\item The concatenation $\mathbf{W} \in \mathbb{R}^{k \times nv}$ of the spatial components of all subjects:
\begin{equation*}
\mathbf{W} = 
\begin{bmatrix}
    \mathbf{W}_{1} &  \mathbf{W}_{2} & \cdots & \mathbf{W}_{n}
\end{bmatrix}
\end{equation*}
where $\mathbf{W}_{i} \in \mathbb{R}^{k \times v}$ contains the $k$ spatial components of subject $i$.

\item The concatenation of the weights $\mathbf{S} \in \mathbb{R}^{mt \times k}$ of all runs:
\begin{equation*}
\mathbf{S} = 
\begin{bmatrix}
    \mathbf{S}^{(1)} \\
    \mathbf{S}^{(2)} \\
    \vdots \\
    \mathbf{S}^{(m)}
\end{bmatrix}
\end{equation*}
where $\mathbf{S}^{(s)} \in \mathbb{R}^{t \times k}$ contains the $k$ weights of run $s$ across time.

\end{itemize}

\begin{figure}
\centering
\includegraphics[scale=0.5]{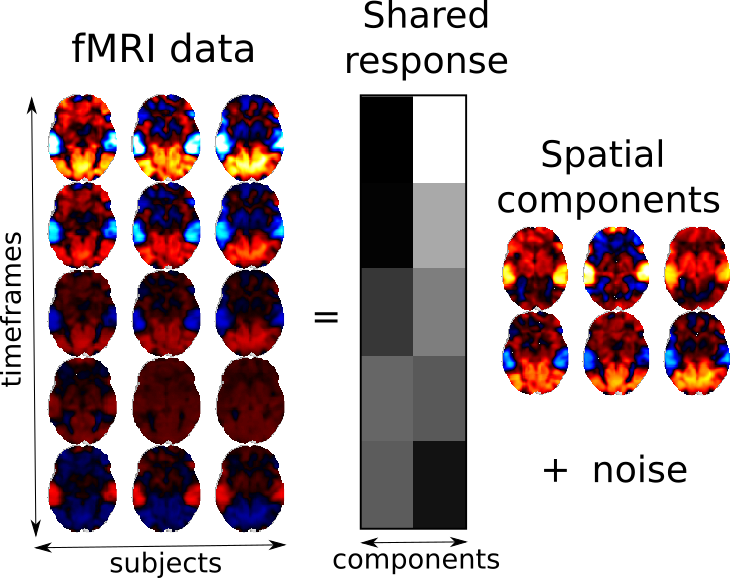}
\caption{\textbf{Shared response model}: The raw fMRI data are modeled as a weighted combination of subject-specific spatial components with additive noise. The weights are shared between subjects and constitute the shared response to the stimuli.}
\label{fig:conceptual_figure1}
\end{figure}

Formally SRM is defined by: 
\[
	\mathbf{X} = \mathbf{S}\mathbf{W} + \mathbf{E},
\]
Where the spatial components $\mathbf{W}$, the shared response $\mathbf{S}$ and the noise $\mathbf{E}$ need to be estimated from $\mathbf{X}$.
An illustration of this definition is given in Figure~\ref{fig:conceptual_figure1}.
Framed this way, group versions of dictionary learning, ICA, blind signal separation or matrix factorization can be seen as particular instances of shared response model.
Most versions of SRM impose orthonormal constrains on spatial components of each subject:
\[
	\forall i \in \{1, \cdots, n\}	\ \mathbf{W}_i \mathbf{W}_i^T = \mathbf{I}_k
\]
where $\mathbf{I}_k \in \mathbb{R}^{k \times k}$ is the identity matrix of size $k$.

\subsection{Deterministic SRM model (DetSRM)}
The deterministic SRM model assumes Gaussian noise with the same variance for all subjects and orthonormal spatial components.
Formally the model reads:
\[
	\forall i \in \{1 \cdots n\}, \ \mathbf{X}_{i}[\tau] \sim \mathcal{N}(\mathbf{S}[\tau] \mathbf{W}_i, \sigma^2 \mathbf{I}_v) \   \text{such that }  \mathbf{W}_i \mathbf{W}_i^T = \mathbf{I}_k,
\]
Maximizing the log-likelihood we obtain the following optimization problem:
\begin{align*}
&\text{min}_{\mathbf{W}_1, \cdots, \mathbf{W}_n, \mathbf{S}} \sum_{i=1}^n ||\mathbf{X}_{i} - \mathbf{S}\mathbf{W}_i||^2 \\
&\text{such that } \forall i \in \{1,\cdots, n\} \  \mathbf{W}_i \mathbf{W}_i^T = \mathbf{I}_k
\end{align*}

This can be solved efficiently using alternate minimization on 
$(\mathbf{W}_1, \cdots , \mathbf{W}_n )$ and $\mathbf{S}$. At each iteration we have two problems to solve that have a closed form solution:
\begin{align*}
	&\forall i \in \{1 \cdots n\}, \\
	& \ \ \text{argmin}_{\{\mathbf{W}_i \text{ such that } \mathbf{W}_i\mathbf{W}_i^T = \mathbf{I}_k\}} \sum_{j=1}^n ||\mathbf{X}_{j} - \mathbf{S}\mathbf{W}_j||^2 = \mathbf{U}_i \mathbf{V}_i,  \\
		   &\ \  \text{where} \ \mathbf{U}_i, \mathbf{D}_i, \mathbf{V}_i = \text{SVD}(\mathbf{S}^T \mathbf{X}_i)
\end{align*}
where SVD stands for singular value decomposition, and
\begin{align*}
&\text{argmin}_{\mathbf{S}} \sum_{i=1}^n ||\mathbf{X}_{i} - \mathbf{S}\mathbf{W}_i||^2 = \frac{1}{n} \sum_{i=1}^n \mathbf{X}_i \mathbf{W}_i^T  \\
\end{align*}

Assuming $k \ll v, t$, the time-complexity of this approach such as implemented in the brainiak library is in $O(n_{iter} nmtvk)$ and storage requirements are in $O(nmvt)$. This means that the method becomes expensive whenever the number of subjects $n$ or runs $m$ becomes large.

\subsection{Probabilistic SRM model (ProbSRM)}
In the probabilistic SRM, spatial components are assumed orthonormal, the shared response is modeled by its covariance matrix $\mathbf{\Sigma} \in \mathbb{R}^{k \times k}$ and the variance of the Gaussian noise $\sigma_1, \cdots , \sigma_n$ is assumed different for different subjects.
In the original paper \cite{chen2015reduced}, an intercept is also learned but since we remove the mean of each time-course as a preprocessing step, it is of no use here.
Formally the model reads:
\begin{align*}
&\mathbf{S}[\tau] \sim \mathcal{N} (0_k, \mathbf{\Sigma}) \\
& \forall i \in \{1 \cdots n\}, \\
& \ \ \mathbf{X}_i[\tau] \sim \mathcal{N} (\mathbf{S}[\tau]\mathbf{W}_i, \sigma_i^2 \mathbf{I}_v) \\
& \ \ \text{such that }  \  \mathbf{W}_i\mathbf{W}_i^T = \mathbf{I}_k
\end{align*}

The optimization is done using an expectation maximization algorithm described in \cite{anderson2016enabling} and \cite{chen2015reduced}. Assuming $k \ll v, t$, the time-complexity of this approach such as implemented in the brainiak library is $O(n_{iter} k nv tm)$ and storage requirements are in $O(nmvt)$. So ProbSRM is also expensive whenever the number of subjects $n$ or runs $m$ becomes large.

\subsection{FastSRM model}
When we deal with large datasets ($n, m, v, t$ are large), above implementations require huge computational power. We introduce FastSRM, a fast and memory-efficient algorithm.
In a first step, we project the data $\mathbf{X} \in \mathbb{R}^{tm \times nv}$ onto a chosen atlas $\mathbf{A}$ with $c$ parcels.
The atlas $\mathbf{A} \in \mathbb{R}^{c \times v}$ can be either probabilistic or a strict partition of the set of voxels but the number of parcels $c$ of the atlas should be larger than the number of components $k$ of the FastSRM model and small compared to the number $v$ of voxels.
In typical settings $c$ reaches hundreds to one thousand.
Projection onto the atlas yields reduced data $\hat{\mathbf{X}} \in \mathbb{R}^{tm \times nc}$.

\[
\forall i \in \{1 \cdots n\}, s \in \{1 \cdots m \} \ \hat{\mathbf{X}}^{(s)}_i = \mathbf{X}^{(s)}_i \mathbf{A}^T (\mathbf{A} \mathbf{A}^T)^{-1}
\]

In cases where the atlas is a partition, projecting onto the atlas is equivalent to averaging brain activation in each parcel of the atlas. 
Formally, denoting $R_j$ the parcel $j$ of atlas $\mathbf{A}$, we compute the projection over the atlas by: 
\[
\forall i \in \{1 \cdots n\}, s \in \{1 \cdots m \} \ \hat{\mathbf{X}}^{(s)}_i[\tau, j] = \frac{\sum_{x \in R_j} \mathbf{X}^{(s)}_i[\tau, x]}{|R_j|}
\]

In a second step we apply our preferred SRM algorithm on the reduced data to find the shared response in reduced space $\hat{\mathbf{S}}$ (in our implementation we use a deterministic SRM). Since $c$ is small compared to $v$, this step is very fast even if the number of iterations is high.
 
\[
	\hat{\mathbf{W}}, \hat{\mathbf{S}} =\text{DetSRM}(\hat{\mathbf{X}})
\]
where the reduced maps $\hat{\mathbf{W}}$ and the shared response in reduced space $\hat{\mathbf{S}}$ are output of the deterministic SRM algorithm (DetSRM).
The spatial components of each subject are recovered by orthonormal regression using the shared response in reduced space $\hat{\mathbf{S}}$ and the data $\mathbf{X}$:
\begin{align*}
&\forall i, \mathbf{W}_i= \mathbf{U}_i \mathbf{V}_i  \\
& \text{where} \ \mathbf{U}_i, \mathbf{D}_i, \mathbf{V}_i = \text{SVD} \left(\sum_{s=1}^m \hat{\mathbf{S}}^{(s)^T} \mathbf{X}^{(s)}_i \right)
\end{align*}

In FastSRM as well as in ProbSRM and DetSRM, fitting the algorithm only means learning the spatial components. If a temporal model is needed it has to be recomputed a posteriori.
In order to compute the shared response from fMRI data of $n$ subjects $\mathbf{X}^{(s)}_1 \cdots \mathbf{X}^{(s)}_n$ in a particular run $s$, one has just to average the projection of the data onto the basis of each subject:
\[
	\mathbf{S}^{(s)} = \frac{1}{n}\sum_{i=1}^n \mathbf{X}^{(s)}_i \mathbf{W}_i^T
\]

Note that when the shared response of train data is needed one cannot use directly the FastSRM shared response  $\hat{\mathbf{S}}$ in reduced space, since it does not have the right scale.
However, the spatial components obtained by orthonormal regression from the ill-scaled shared response are valid. 
Indeed if we multiply the shared response by a scaling factor $f$, for any subject $i$ the singular value decomposition of $\sum_{s=1}^m f \mathbf{\hat{S}}^{(s)^T} \hat{\mathbf{X}}^{(s)}_i$ is given by $\mathbf{U}_i f \mathbf{D}_i \mathbf{V}_i$ where $\mathbf{U}_i \mathbf{D}_i \mathbf{V}_i$ is the singular decomposition we would obtain with $f=1$ and therefore the spatial components of subject $i$ are given by $\mathbf{W}_i = \mathbf{U}_i \mathbf{V}_i$ which is independent from the scale factor.

The orthonormal projection can easily be replaced by any kind of regression. One can easily impose sparsity, non-negativity, smoothness or similarity constrains on spatial components instead of orthonormality to obtain more interpretable patterns.
This however leads to more heavier computations.
In our implementation, we use orthonormal regression.
Assuming $k \ll c \ll v, t$, the bottleneck is the projection onto the atlas which yields a time complexity in $O(n m t v c)$.
We need to keep spatial components, the atlas as well as one run of one subject in memory which yields a memory complexity of $O(nvk + vc + vt))$.
If parallelization is used memory requirements become $O(nvk + vc + n_{jobs} vt))$ and fitting time becomes $O(\frac{nmtvc}{n_{jobs}})$ where $n_{jobs}$ is the number of jobs. 
It is also possible to write spatial components on disk instead of keeping them in memory which adds about $O(vk)$ read/write operations but reduces the memory complexity to $O(v(c + n_{jobs}t))$. In our experiments we write spatial components on the disk.

\begin{figure}
\centering
\includegraphics[scale=0.34]{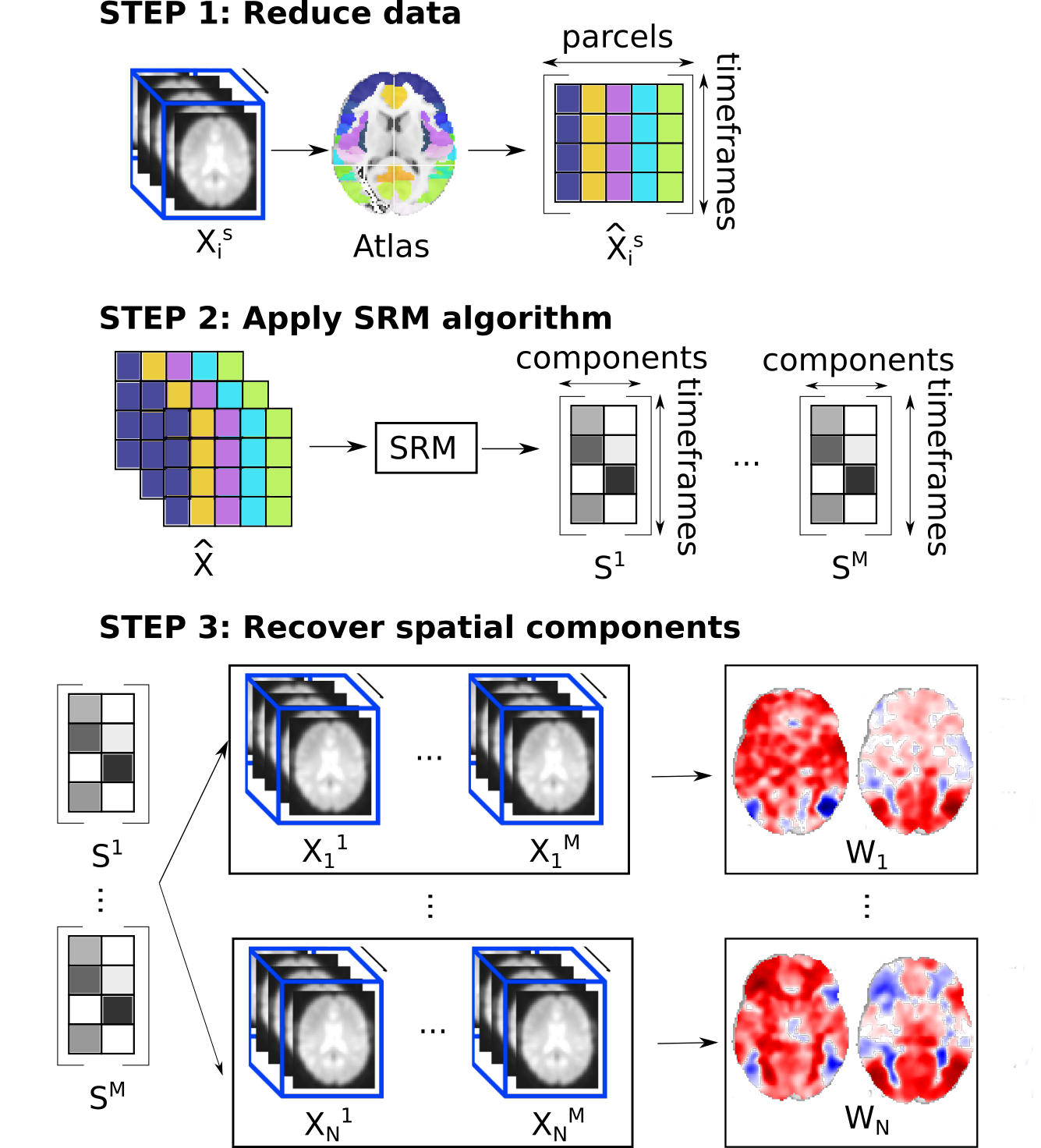}
\caption{\textbf{FastSRM algorithm} In step 1, data are projected onto an atlas (top). In step 2 a deterministic SRM algorithm is applied on reduced data to compute the shared response(middle). In step 3, spatial components are recovered by regression from the shared response (bottom).}
\label{fig:conceptual_figure2}
\end{figure}

\subsection{Experiments}
\subsubsection{Datasets}

We use five fMRI datasets of subjects exposed to naturalistic stimuli. When needed, datasets are preprocessed with FSL \url{http://fsl.fmrib.ox.ac.uk/fsl} using slice time correction, spatial realignment, coregistration to the T1 image and affine transformation of the functional volumes to a template brain (MNI). Using nilearn \cite{abraham2014machine}, preprocessed data are resampled, masked (using a full brain mask available at \url{http://cogspaces.github.io/assets/data/hcp_mask.nii.gz}), detrended and standardized after a 5 mm smoothing is applied.  
\paragraph{SHERLOCK}
In SHERLOCK dataset, 17 participants are watching "Sherlock" BBC TV show (episode 1). 
These data are downloaded from \url{http://arks.princeton.edu/ark:/88435/dsp01nz8062179}. 
Data were acquired using a 3T scanner with an isotropic spatial resolution of 3 mm. 
More information including the preprocessing pipeline is available in \cite{sherlock}.
Subject 5 is removed because of missing data leaving us with 16 participants.
Although SHERLOCK data contains originally only 1 run, we split it into 4 runs of 395 timeframes and one run of 396 timeframes for the needs of our experiments.

\paragraph{FORREST}
In FORREST dataset 20 participants are listening to an audio version of the movie Forrest Gump.
FORREST data are downloaded from OpenfMRI \cite{poldrack2013toward}. 
Data were acquired using a 7T scanner with an isotropic spatial resolution of 1 mm (see more details in \cite{hanke2014high}.
More information about the forrest project can be found at \url{http://studyforrest.org}.
Subject 10 and run 8 are discarded because of missing data.
We therefore use full brain data of 19 subjects split in 7 runs of respectively 451, 441, 438, 488, 462, 439 and 542 timeframes.
 
\paragraph{RAIDERS}
In RAIDERS dataset, 10 participants are watching the movie "Raiders of the lost ark".
The RAIDERS dataset pertains to the Individual Brain Charting dataset (\cite{ibc}).
They acquired at NeuroSpin using a 3T scanner with an isotropic spatial resolution of 3 mm.
The RAIDERS dataset reproduces the protocol described in \cite{haxby2011common}.
We use full brain data of 10 subjects split in 9 runs of respectively 374, 297, 314, 379, 347, 346, 350, 353 and 211 timeframes.

\paragraph{CLIPS}
In CLIPS dataset, 10 participants are exposed to short clips. 
The CLIPS dataset also pertains to the Individual Brain Charting dataset (\cite{ibc}).
It reproduces the protocol of original studies described in \cite{nishimoto2011reconstructing} and \cite{huth2012continuous}.
In our experiments we use the data of 10 participants acquired in 17 runs of 325 timeframes.

At the time of writing, the CLIPS and RAIDERS dataset from the individual brain charting dataset \url{https://project.inria.fr/IBC/} are not yet public, but they will be in the future. Protocols on the visual stimuli presented are available in a dedicated repository on Github: \url{https://github.com/hbp-brain-charting/public_protocols}. The informed consent of all subjects was obtained before scanning.

\paragraph{CamCAN}
In CamCAN dataset, 647 participants aged from 18 to 88 years are watching Alfred Hitchcock's "Bang! You're Dead" (edited so that it lasts only 8 minutes).
CamCAN consists of data obtained from the CamCAN repository (available at \url{http://www.mrc-cbu.cam.ac.uk/datasets/camcan/}) (see \cite{taylor2017cambridge} and \cite{shafto2014cambridge}).
We use all available subjects and runs yielding 647 participants and 1 run of 193 timeframes.

A summary about the size of each dataset is available in Table~\ref{tab:dataset_desc}.

\subsubsection{fMRI reconstruction: Evaluate the ability to recover BOLD signal on left-out runs}
\label{reconstruction}

When subjects are exposed to the same stimuli, SRM algorithms posit that the recorded fMRI data can be modeled as a product of two matrices, one of which is fixed across time but subject-specific (the spatial components) while the other varies across time but is common to all subjects (the shared response).
Under this framework, once spatial components are known, we can generate an accurate estimation of the data of one subject given the data of all others.
In this experiment we test whether we can recover data of a left-out subject using previous data of the same subject as well as data from other subjects. 
We denote $\mathbf{X}^{(-s)}$ brain recordings of all runs but run $s$:
 \begin{equation*}
	 \mathbf{X}^{(-s)} = 
\begin{bmatrix}
	\mathbf{X}^{(1)} \\
	\vdots \\
	\mathbf{X}^{(s-1)} \\
	\mathbf{X}^{(s+1)} \\
	\vdots \\
	\mathbf{X}^{(m)}
\end{bmatrix}
\end{equation*}

Similarly, $\mathbf{X}_i^{(-s)}$ refers to the brain recordings of subject $i$ using all runs but run $s$. The data from all subjects but $i$ acquired during run $s$ are denoted:
\begin{equation*}
	\mathbf{X}^{(s)}_{-i} = 
\begin{bmatrix}
	\mathbf{X}_1^{(s)} & \cdots & \mathbf{X}^{(s)}_{i-1} & \mathbf{X}^{(s)}_{i+1} & \cdots & \mathbf{X}^{(s)}_{n} 
\end{bmatrix}
\end{equation*}

We evaluate our model using a cross validation scheme known as co-smoothing (see \cite{wu2018learning}). First, all brain recordings for all runs but one $\mathbf{X}^{(-s)}$ are used for learning subjects spatial components $\mathbf{W}^{(-s)}$. The exponent in $\mathbf{W}^{(-s)}$ indicates that these spatial components were learned using all runs but run $s$. 

\begin{equation*}
	\mathbf{W}^{(-s)}, \mathbf{S}^{(-s)} = SRM(\mathbf{X}^{(-s)})
\end{equation*}

Then we focus on the left-out run $s$ and use all subjects but one $\mathbf{X}^{(s)}_{-i}$ to compute a shared response for the left-out run $\mathbf{S}^{(s)}$.

\begin{equation*}
	\mathbf{S}^{(s)} = \frac{\sum_{z=1, z\neq i}^n \mathbf{X}^{(s)}_z\mathbf{W}^{(-s)^T}_z}{n - 1}
\end{equation*}

From $\mathbf{S}^{(s)}$ and $\mathbf{W}_i^{-s}$ we compute $\widetilde{\mathbf{X}}^{(s)}_i$ which stands as an estimate of brain activity of the left-out subject $i$ during the left-out run $s$.

\begin{equation*}
	\widetilde{\mathbf{X}}_i^{(s)} = \mathbf{W}^{(-s)}_i \mathbf{S}^{(s)}
\end{equation*}

\begin{figure}
\centering
\includegraphics[scale=0.24]{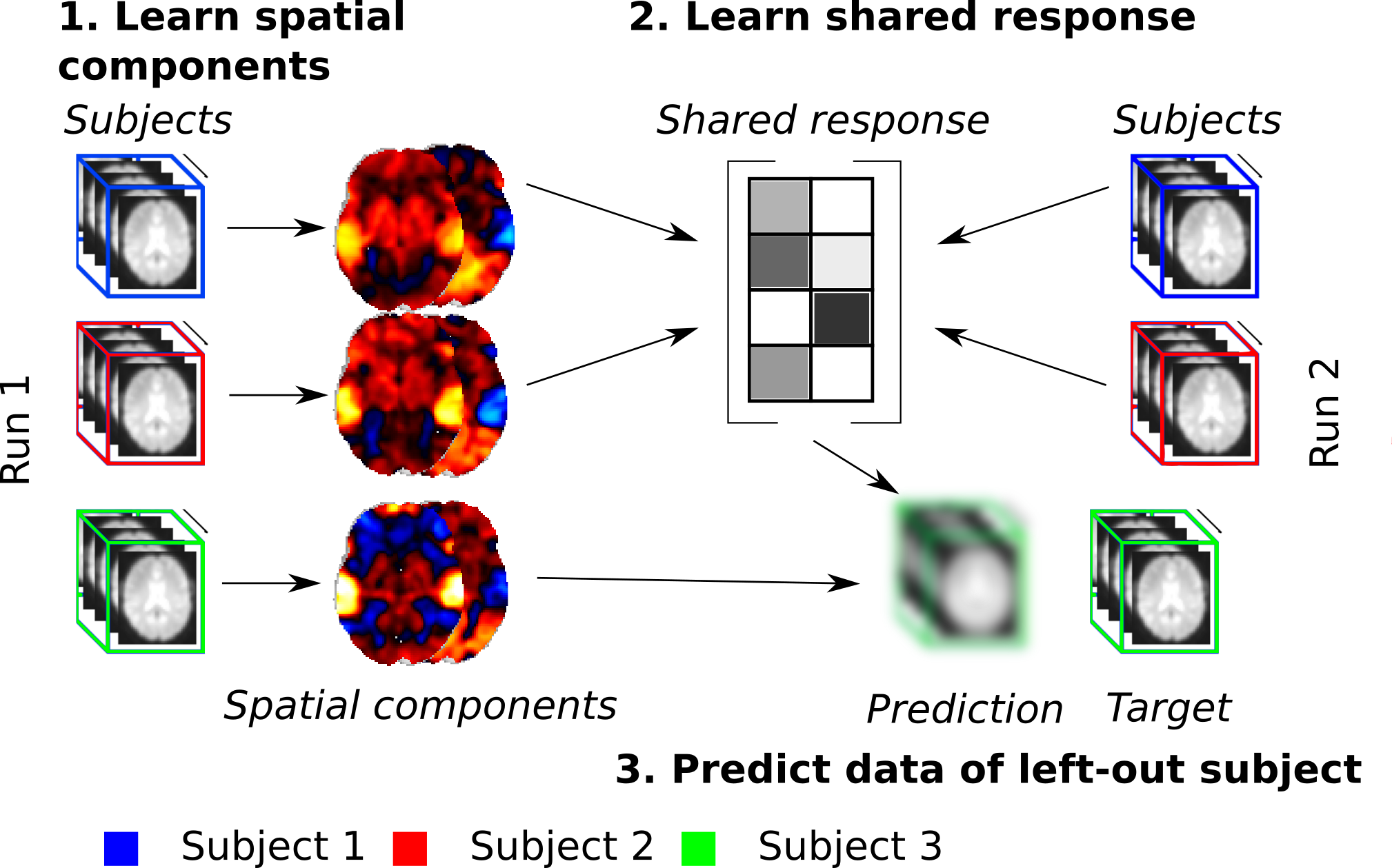}
\caption{\textbf{Experiment — Reconstruct data from a left-out subject} All runs but one are used to compute spatial components for every subject (left).
  Then spatial components and data from the left-out run of all subjects but one are used to compute the shared response in the left-out run.
  At last, the shared response during the left-out run and  the spatial components of the test subject are used to predict the data of the test subject in the left-out run.
  The performance of the model is measured by comparing the prediction and true data using the $R^2$ score.}
\label{fig:experiment_reconstruction}
\end{figure}

An illustration of our reconstruction experiment is available in Figure~\ref{fig:experiment_reconstruction}.
The performance is measured voxel-wise using the $R^2$ score between $\widetilde{\mathbf{X}}_i^{(s)}$ and $\mathbf{X}_i^{(s)}$ as a similarity measure.
For any two time-courses $x \in \mathbb{R}^t$ and $y \in \mathbb{R}^t$ we define the $R^2$ score by:

\[
	R^2(x, y) = 1 - \frac{\sum_{z=1}^t (x[z] - y[z])^2}{\sum_{z=1}^t (y[z] - \overline{y}[z])^2} 
\]

Where $\overline{y} = \frac{1}{t}\sum_{z=1}^t y[z]$.
Following the leave-one-out cross validation scheme, all our experiments are done several times with a different left-out subject to reconstruct. 
We measure the average $R^2$ score across  all left-out subjects. Note that we obtain one such value per voxel.

\subsubsection{Predict age from spatial components}
\begin{figure}
\centering
\includegraphics[scale=0.35]{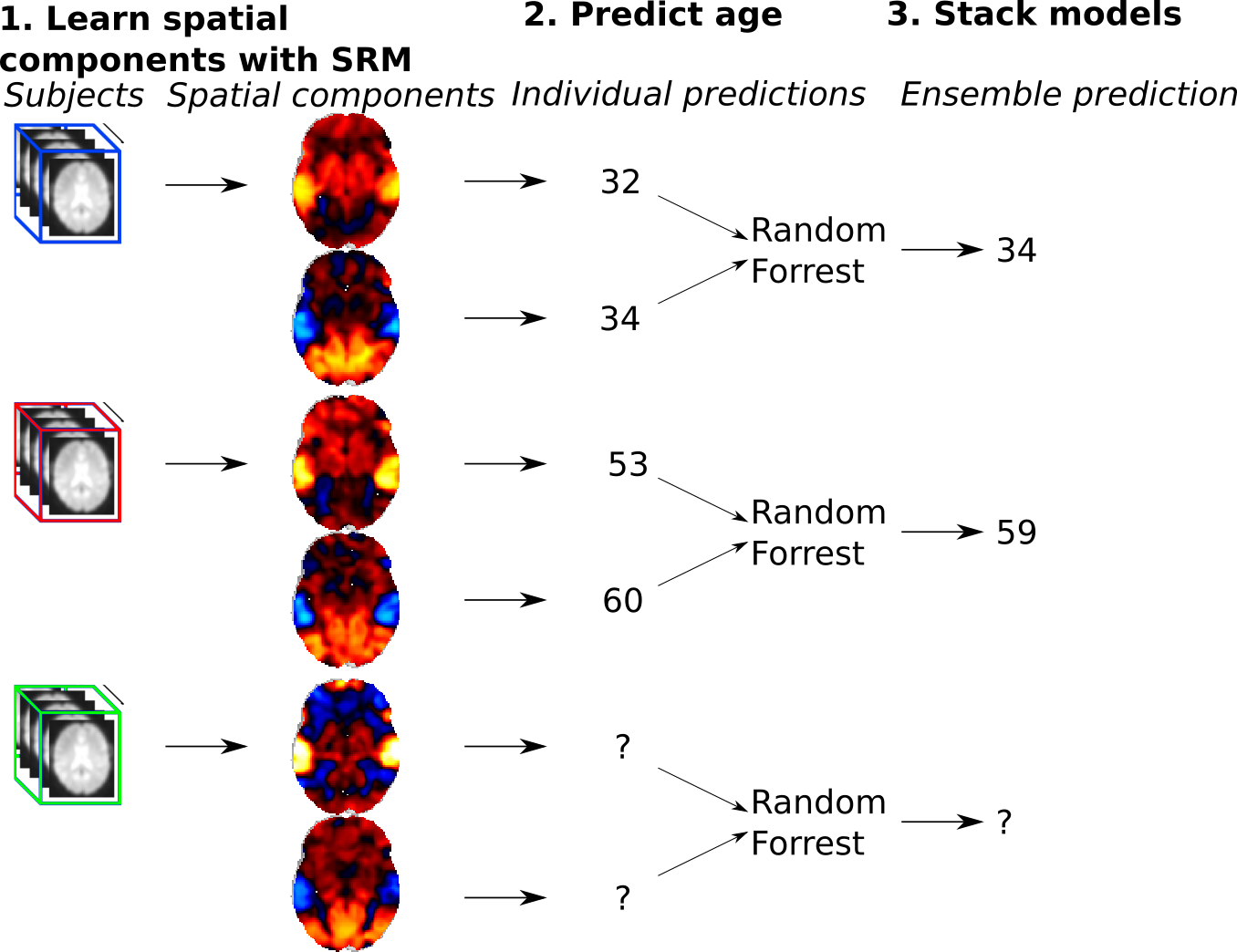}
\caption{\textbf{Experiment — Predict age from spatial components extracted using FastSRM}: We first learn the spatial components from fMRI data using SRM. We learn one Ridge model per spatial components to predict age across subjects. Then, these models are aggregated using a Random Forest (like in \cite{rahim2017joint}).} 
\label{fig:experiment_age_prediction}
\end{figure}

Since spatial components are subject-specific they should be predictive of subject-specific features such as age.
In this experiment we try to predict subject's age from movie-watching data using SRM algorithms.
Functionally matched spatial components are obtained using an SRM algorithm.
They are divided into two groups (train and test data) where the train set contains 80\% of the data and the test set 20\%.
Within the train set we split again our data into two groups: the first group is used to train one Ridge model per spatial components, the second group is used to train a Random Forest to predict age from Ridge predictions. This way of stacking models is similar to the pipeline used in \cite{rahim2017joint}.
We use 5 fold cross validation to split the train set (so that the number of samples used to train the Random Forest is the number of elements in the train set).
Then the train set is used to train one Ridge model per spatial component.
On the test set each Ridge model makes a prediction and the predictions are aggregated using the Random Forest model.
An illustration of the process is available in Figure~\ref{fig:experiment_age_prediction}. 

In each Ridge model, the coefficient that determines the level of l2 penalization is set by generalized cross validation, an efficient form of leave-one-out cross validation (see the RidgeCV implementation of Scikit Learn \cite{pedregosa2011scikit}).

The train and test sets are chosen randomly. 5 different choices for the train and test set are made. We report the average mean absolute error (MAE) on the test set averaged over the 5 splits.

\section{Results and Discussion}

\subsection{fMRI Reconstruction}
We perform the reconstruction experiment on the FORREST, CLIPS, RAIDERS and SHERLOCK datasets. We compare brainiak's implementation of ProbSRM / DetSRM to our implementation of FastSRM in terms of fitting time, memory usage and performance. In order to be fair, we do not use parallelization ($n_{jobs} = 1$) and we set the number of iterations to 10 ($n_{iter}=10$) which is ProbSRM's default. Note that the fitting time of ProbSRM is roughly proportional to the number of iterations while it has a limited impact on the fitting time of FastSRM.

We run our experiments on the full brain and report an $R^2$ score per voxel.
However we measure the performance in terms of mean $R^2$ score inside a region of interest (in order to leave out regions where there is no useful information).
In order to determine the region of interest, we focus on the results of ProbSRM with $10, 20, 50$ and $100$ components and keep only the intersection of regions where the $R^2$ score is above $0.05$.
This means of selecting regions favors ProbSRM. For completeness, full brain $R^2$ images obtained on the four datasets with ProbSRM and FastSRM using $20$ components averaged across subjects are available in Figure~\ref{fig:example_r2}.

\begin{figure}
\centering
\includegraphics[scale=0.34]{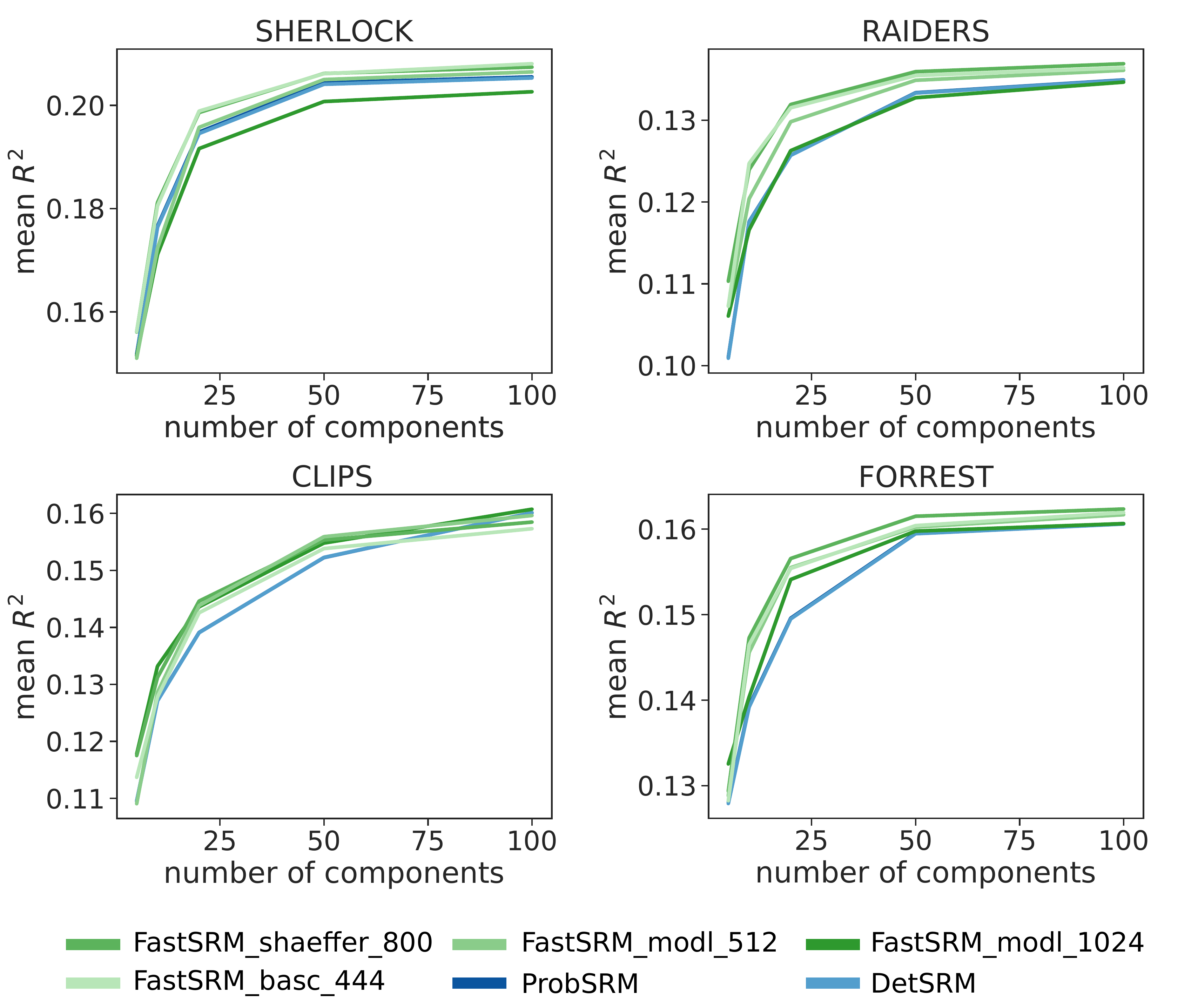}
\caption{\textbf{Performance of the methods in an encoding test} We compare the performance (measured in terms of average $R^2$ score in a region of interest) of ProbSRM and FastSRM with different atlases in function of the number of components used. Atlases tested are MODL with 512 and 1024 parcels, Basc with 444 parcels and Shaeffer with 800 parcels. Datasets tested are SHERLOCK (top left), RAIDERS (top right), CLIPS (bottom left) and FORREST (bottom right). 
As we can see, no matter which atlas is chosen, FastSRM matches ProbSRM's performance.}
\label{fig:mean_r2_score}
\end{figure}

In Figure~\ref{fig:mean_r2_score}, we plotted the mean $R^2$ score against the number of components ($k$) for ProbSRM, DetSRM and FastSRM algorithm with different atlases.
The $R^2$ score tends to increase with the number of components (which is what is expected as more information can be retrieved when the number of components is high).
FastSRM matches the performance of ProbSRM and DetSRM. This holds for any atlas we chose (BASC (444 parcels), SHAEFFER (800 parcels), MODL (512 and 1024 parcels)) and for all datasets we tried (SHERLOCK, RAIDERS, CLIPS and FORREST). 

\begin{figure}
\centering
\includegraphics[scale=0.33]{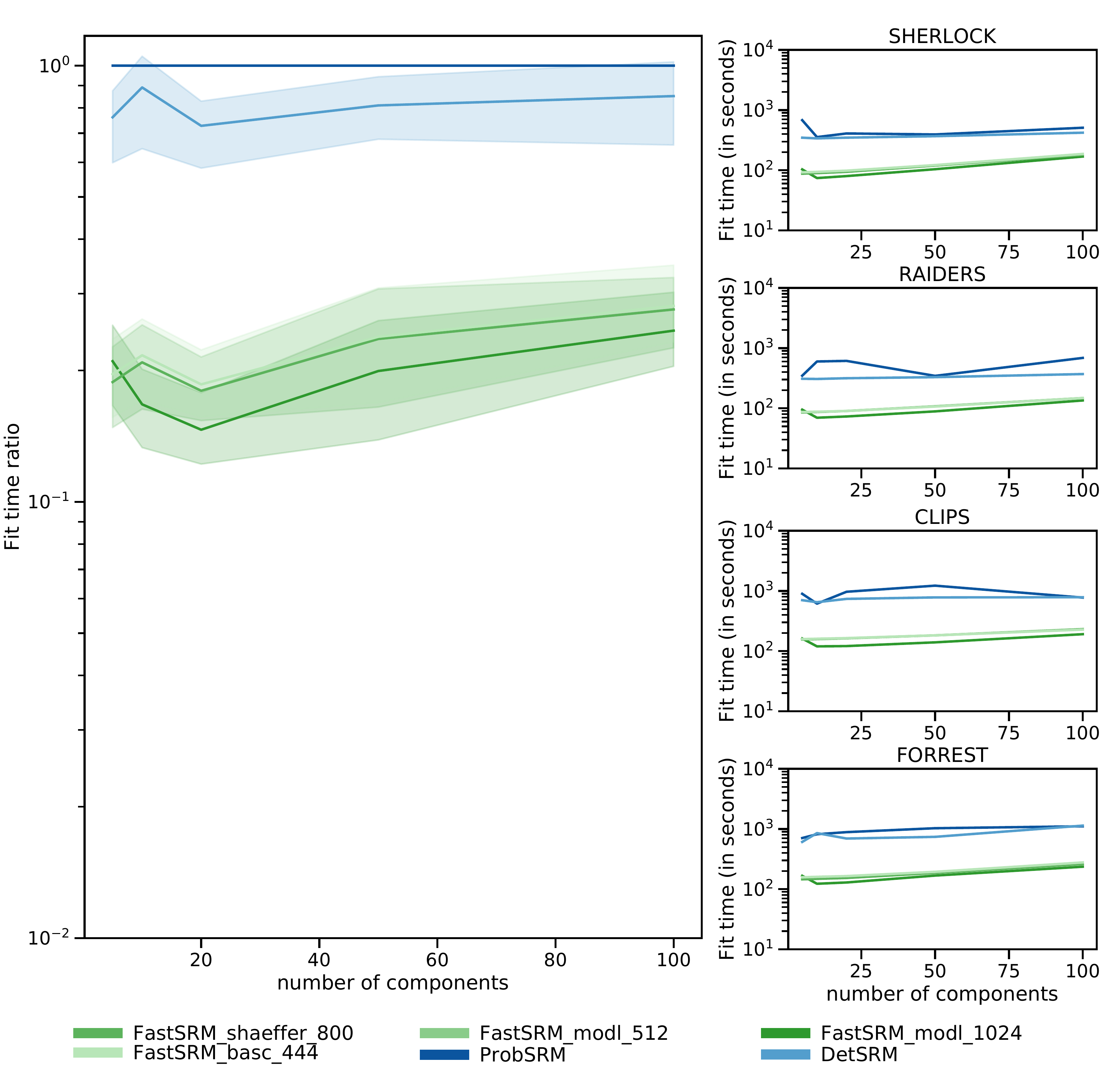}

\caption{\textbf{Fitting time of FastSRM, ProbSRM and DetSRM} We compare the fitting time of ProbSRM, DetSRM and FastSRM with different atlases in function of the number of components used. Atlases tested are MODL with 512 and 1024 parcels, Basc with 444 parcels and Shaeffer with 800 parcels. Datasets tested are SHERLOCK, RAIDERS, CLIPS and FORREST.
\textbf{Left}: Fitting time (as a fraction of ProbSRM fitting time) averaged over the four datasets.
\textbf{Right}: Fitting time (in seconds) for each of the four different datasets.
FastSRM is about 5 times faster than ProbSRM.}
\label{fig:fit_time}
\end{figure}

In Figure~\ref{fig:fit_time}, we compare the running time of FastSRM, DetSRM and ProbSRM on the four different datasets.
FastSRM is on average (across datasets) about 5 times faster.
On the FORREST dataset, we compute a shared response in about 3 minutes when it takes about 20 minutes with ProbSRM or DetSRM.

In Figure~\ref{fig:memory_usage}, we compare the memory (RAM) consumption of FastSRM, DetSRM and ProbSRM on the four different datasets.
FastSRM is 20 to 40 times more memory friendly than ProbSRM and 10 to 20 times more memory friendly than DetSRM. On the FORREST dataset the memory usage of FastSRM is between 1 and 3 Go depending on the number of components and the atlas used. Most modern laptops meet these requirements. On the same dataset memory consumption is about 80 Go for ProbSRM and 40 Go for DetSRM which is manageable but costly for small labs.
Overall FastSRM yields same performance as ProbSRM and DetSRM while being much faster and using far less memory. We also show that the atlas used to reduce the data only has a minor impact on performance.

\subsection{Predicting age from spatial components}
Because FastSRM is fast and memory efficient, it enables large-scale analysis of fMRI recordings of subjects exposed to the same naturalistic stimuli.
We use all 647 subjects of the CamCAN dataset and demonstrate the usefulness of FastSRM by showing that the spatial components it extracts from movie watching data are predictive of age.
A key asset of FastSRM is that these spatial components can be visualized and therefore provide meaningful insights. 

\begin{figure}
\centering
\includegraphics[scale=0.44]{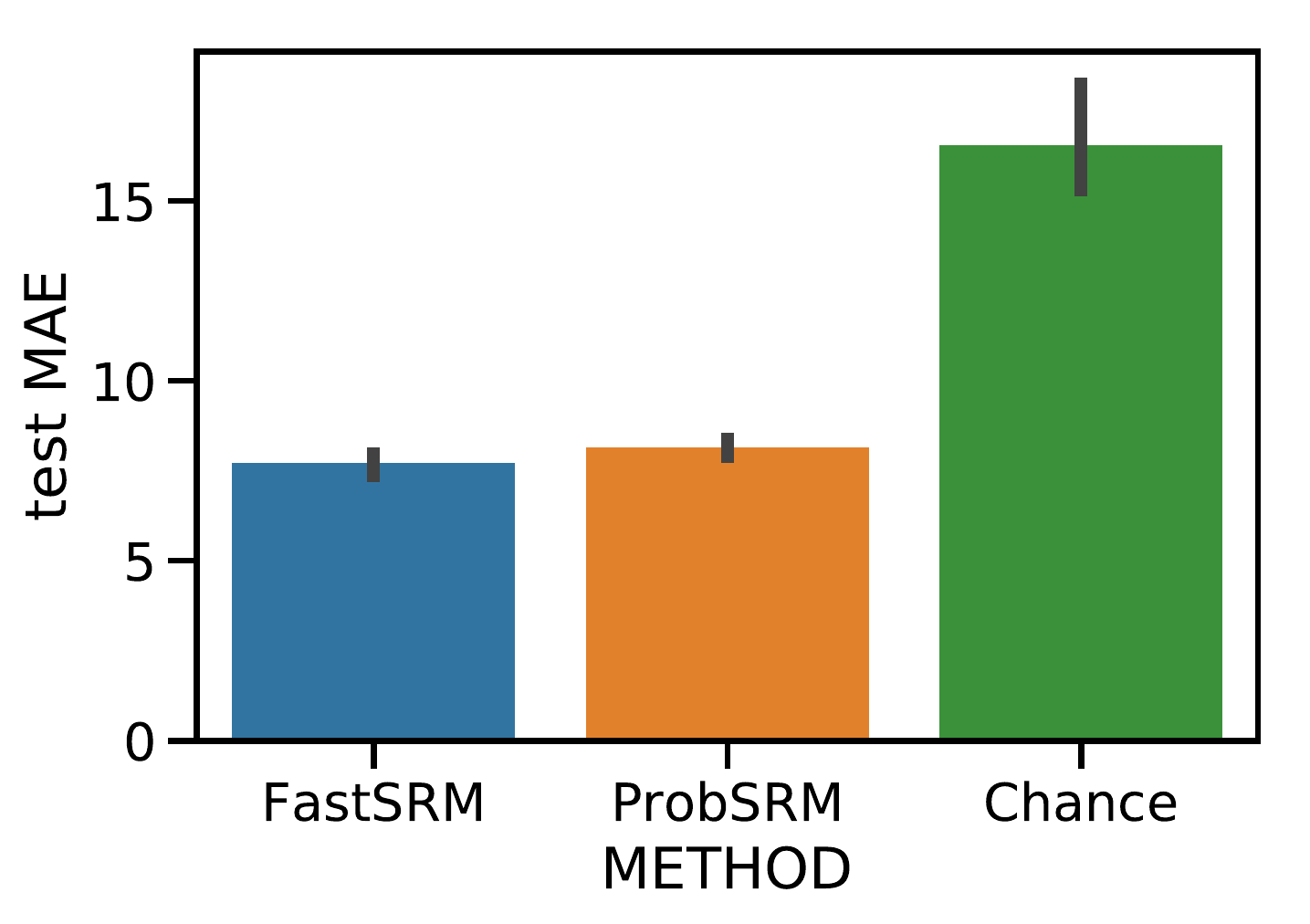}

\begin{tabular}{|c|c|c|}
	\hline
	Algorithm & Memory usage (in Go) & Fitting time (in minutes) \\
	\hline
	FastSRM & 1.1 & 15 \\
	ProbSRM & 186.3 & 69 \\
	\hline
\end{tabular}
\includegraphics[scale=0.365]{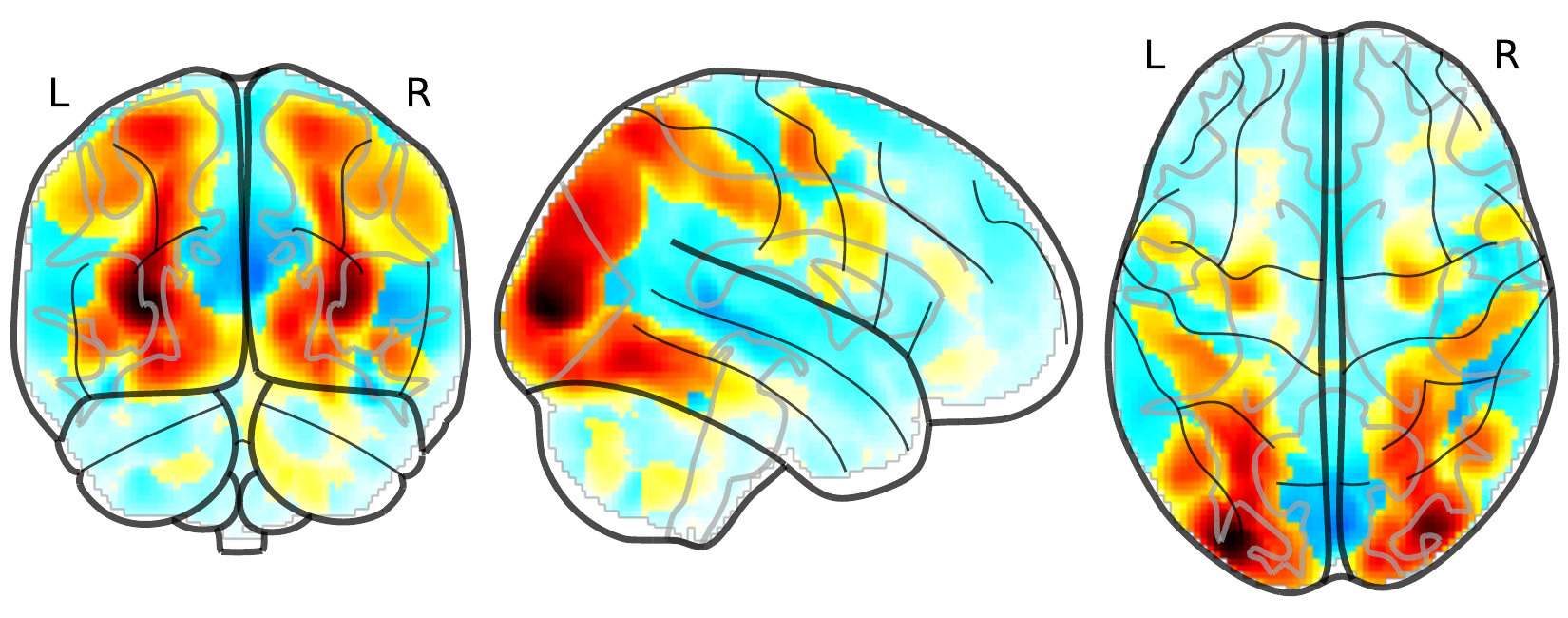}
\includegraphics[scale=0.365]{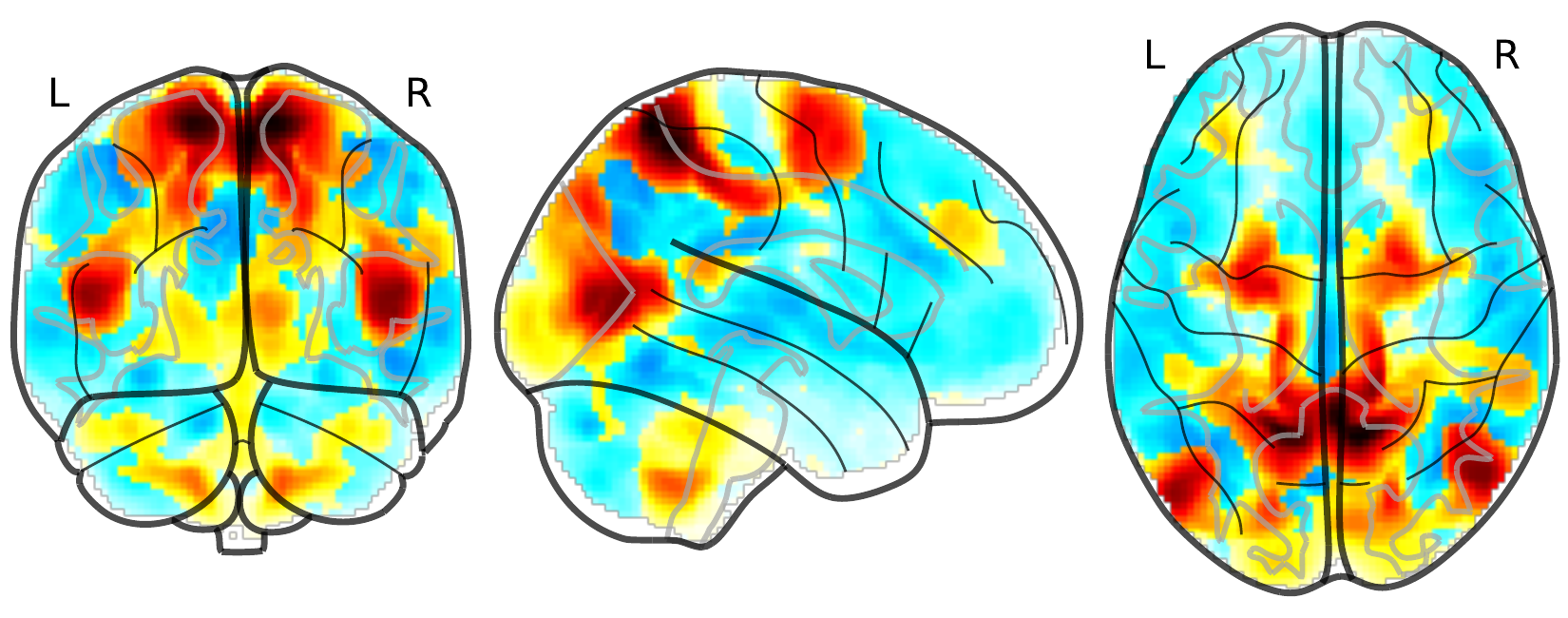}
\includegraphics[scale=0.365]{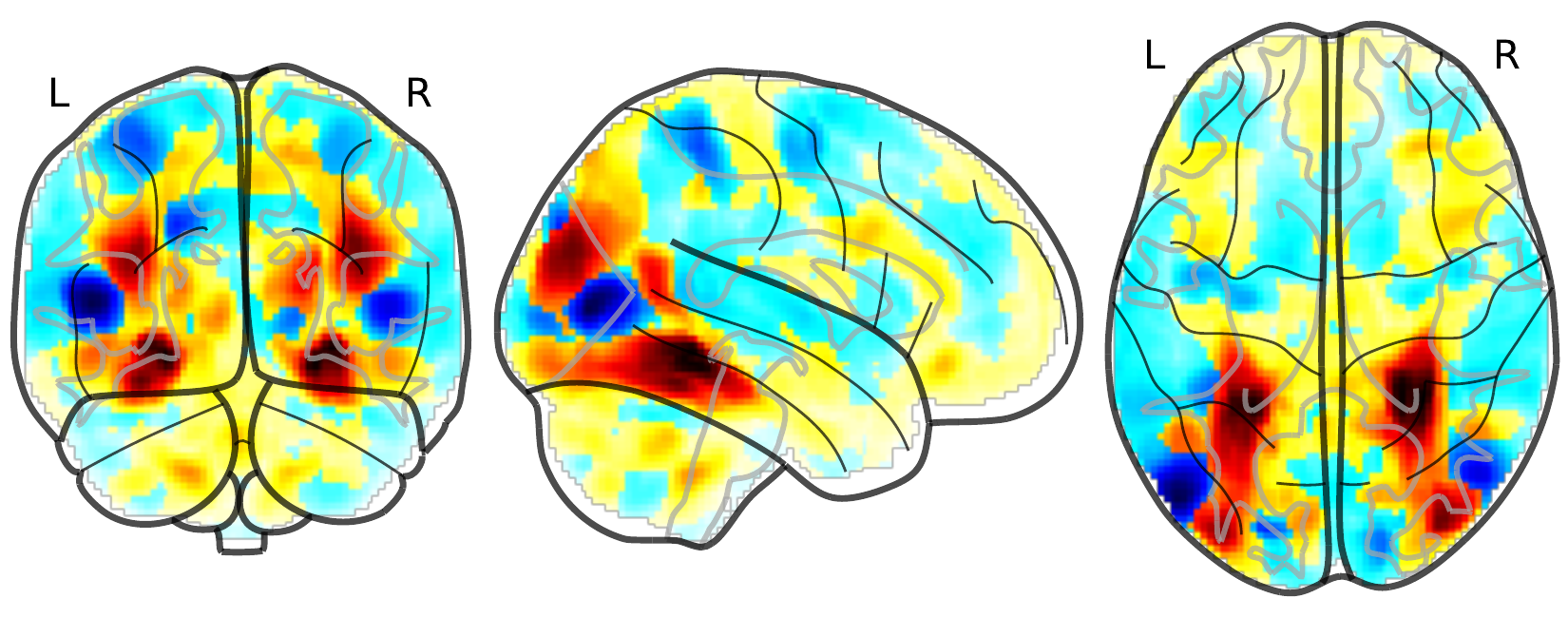}

\caption{\textbf{Age prediction from spatial components}: (top) FastSRM predicts age with a good accuracy (better than ProbSRM and a lot better than chance) resulting in a mean absolute error (MAE) of 7.5 years. (middle) FastSRM is more than 4x faster than ProbSRM and uses 150x less memory, hence it scales better than ProbSRM. (bottom) The three most important spatial components in terms of the reduction in Gini impurity they bring (see Gini importance or Feature importance in \cite{breiman2001random}, \cite{louppe2013understanding}). From top to bottom, the most important spatial component (feature importance: 16\%) highlights the visual dorsal pathway, the second most important spatial component (feature importance: 12\%) highlights the precuneus and the third most important spatial component (feature importance: 8\%) highlights the visual ventral pathway.} 
\label{fig:predict_age}
\end{figure}

Figure~\ref{fig:predict_age} shows that FastSRM predicts age with a good accuracy (better than ProbSRM and a lot better than chance) resulting in a mean absolute error (MAE) of 7.5 years.
It also shows that on CamCAN data, FastSRM is 4x faster and more than 150x more memory efficient than ProbSRM. As before and in order to ensure fair comparison the number of iterations is set to $10$ and we do not make use of parallelization. Note that the memory requirements of ProbSRM on the CamCAN dataset (186Go) make it difficult to use.
FastSRM does not suffer from memory issues, making it suitable to analyse big datasets.

\begin{figure}
\centering
\includegraphics[scale=0.35]{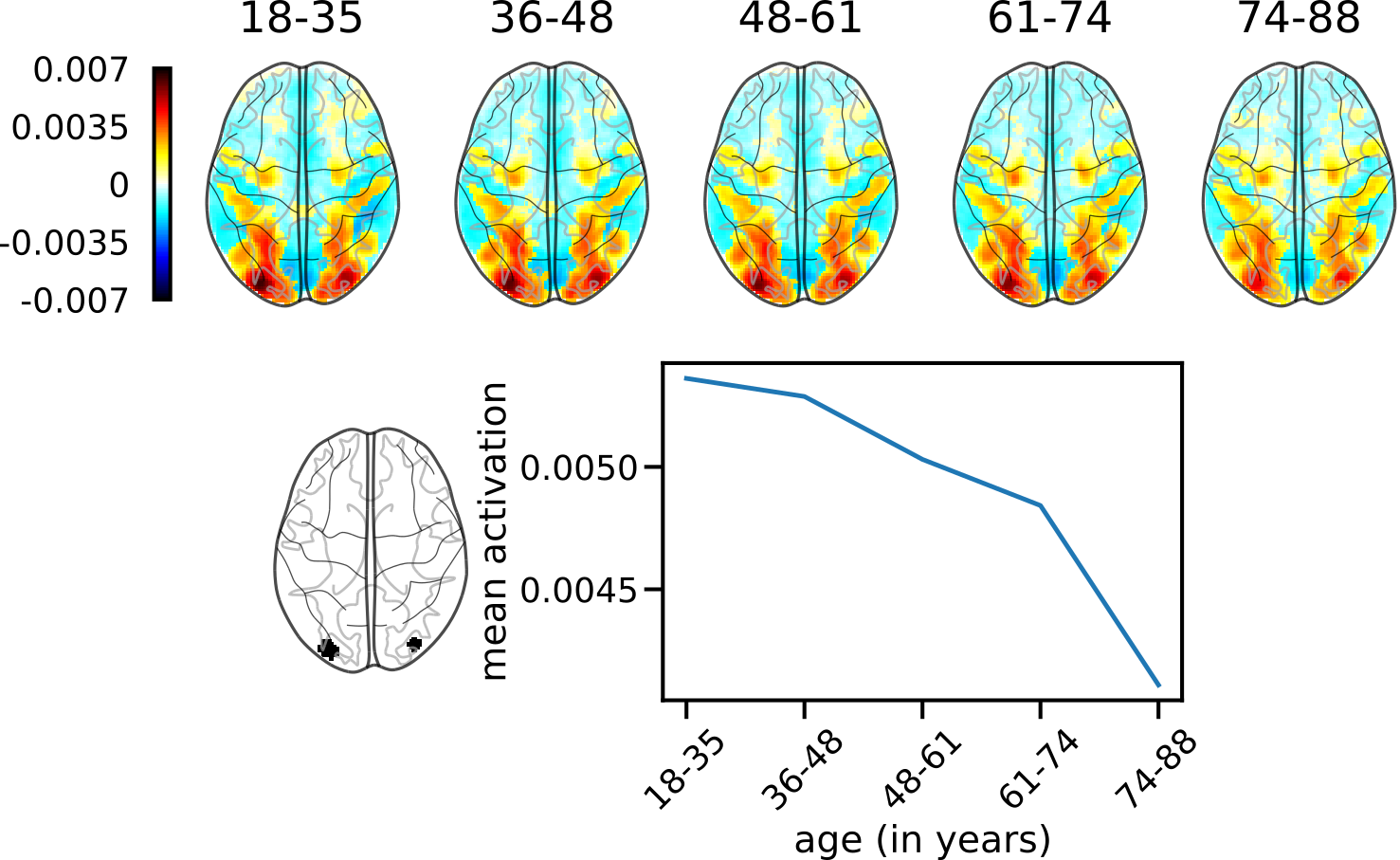}
\caption{\textbf{Evolution of the most predictive spatial component with age}: (Top) Spatial component most predictive of age averaged within groups of different age (18-35, 36-48, 48-61, 61-74, 74-88). (Bottom) Mean activation in the region highlighted by the mask on the left. We see that the activity in the dorsal pathway decreases with age, which explains why this spatial component is a good predictor of age.}
\label{fig:predict_age_interpretation}
\end{figure}

A key asset of our pipeline is that we can see which spatial components are most predictive of age by using feature importance.
Feature importance is assessed by the Gini importance defined in \cite{breiman2001random} or \cite{louppe2013understanding}.
It measures for each feature the relative reduction in Gini impurity brought by this feature.
Feature importance varies with different splits. We use the averaged feature importance over the 5 splits of our pipeline.
In Figure~\ref{fig:predict_age} are shown the 3 most important spatial components representing respectively 16\%, 12\% and 8\% of total feature importance.
These spatial components in decreasing order of importance represent the visual dorsal pathway, the precuneus and the visual ventral pathway. 
The fact that averaged spatial components are interpretable and meaningful allows us to study the influence of age on brain networks involved in movie-watching.
In Figure~\ref{fig:predict_age_interpretation}, we plot the most important spatial component averaged within groups of ages.
We see that these spatial components evolve with age allowing us to visually identify which regions are meaningful. 
It turns out that aging is mostly reflected in brain activity as a
fading of activity in the spatial correlates of movie watching,
particularly in the dorsal visual cortex.
%

\subsection{Conclusion}
As studies using naturalistic stimuli will tend to become more common and large within and across subjects, we need scalable models especially in terms of memory usage.
This is what FastSRM provides.
We show that while FastSRM matches the performance of ProbSRM or DetSRM, it is
significantly faster and requires a lot less memory. While FastSRM's scalability relies on the use of atlases to compress the BOLD signal, we show that the precise choice of the atlas has only marginal effects on performance.

FastSRM allows large scale analysis of fMRI data of subjects exposed to naturalistic stimuli. As one example of such analysis, we show that it can be used to predict age from movie-watching data. Interestingly, although FastSRM is an unsupervised model, it extracts meaningful networks and as such constitutes a practical way of studying subjects exposed to naturalistic stimuli.

We also show that individual information can be extracted from the fMRI activity when subjects are exposed to naturalistic stimuli. Our predictive model is reminiscent of that of \cite{bijsterbosch2018relationship}, that have shown that ICA components obtained from the decomposition of resting state data carry important information on individual characteristics. 

As a side note, we chose to keep the orthonormality assumptions of the original SRM model but slight modifications of our implementation of FastSRM would allow one to build more refined model promoting sparsity, non-negativity or smoothness of spatial components for example.

The remaining difficulty with SRM is to interpret the spatio-temporal decomposition. Reverse correlation \cite{hasson2004intersubject} can be used to clarify the cognitive information captured in the shared response.

Our code is freely available at \url{https://github.com/hugorichard/brainiak/tree/fastsrm}.

\appendix

\section{Appendices}
\begin{figure}
\centering
\includegraphics[scale=0.33]{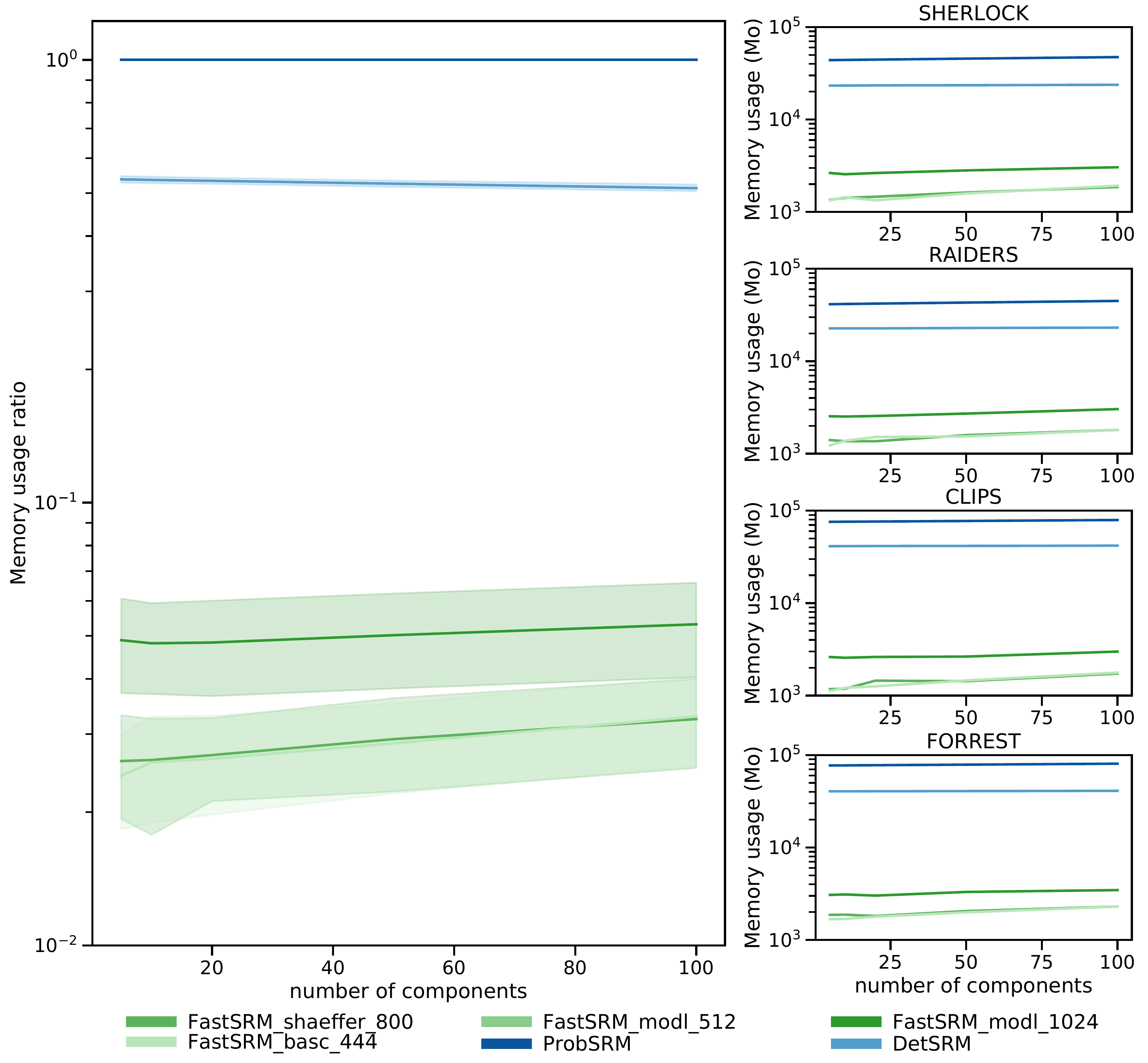}
\caption{\textbf{Memory usage of FastSRM, ProbSRM and DetSRM} We compare the memory usage of DetSRM, ProbSRM and FastSRM with different atlases in function of the number of components used. Atlases tested are MODL with 512 and 1024 parcels, Basc with 444 parcels and Shaeffer with 800 parcels. Datasets tested are SHERLOCK, RAIDERS, CLIPS and FORREST.
\textbf{Left}: Memory usage (as a fraction of ProbSRM memory usage) averaged over the four datasets
\textbf{Right}: Memory usage (in Mo) for each of the four different datasets.
FastSRM with probabilistic atlases (MODL) is about 20x more memory efficient than ProbSRM and 40x with deterministic atlases (Basc, Shaeffer) making it possible to compute a shared response on a large dataset using a modern laptop.}
\label{fig:memory_usage}
\end{figure}

\begin{figure}
\centering
\includegraphics[scale=0.43]{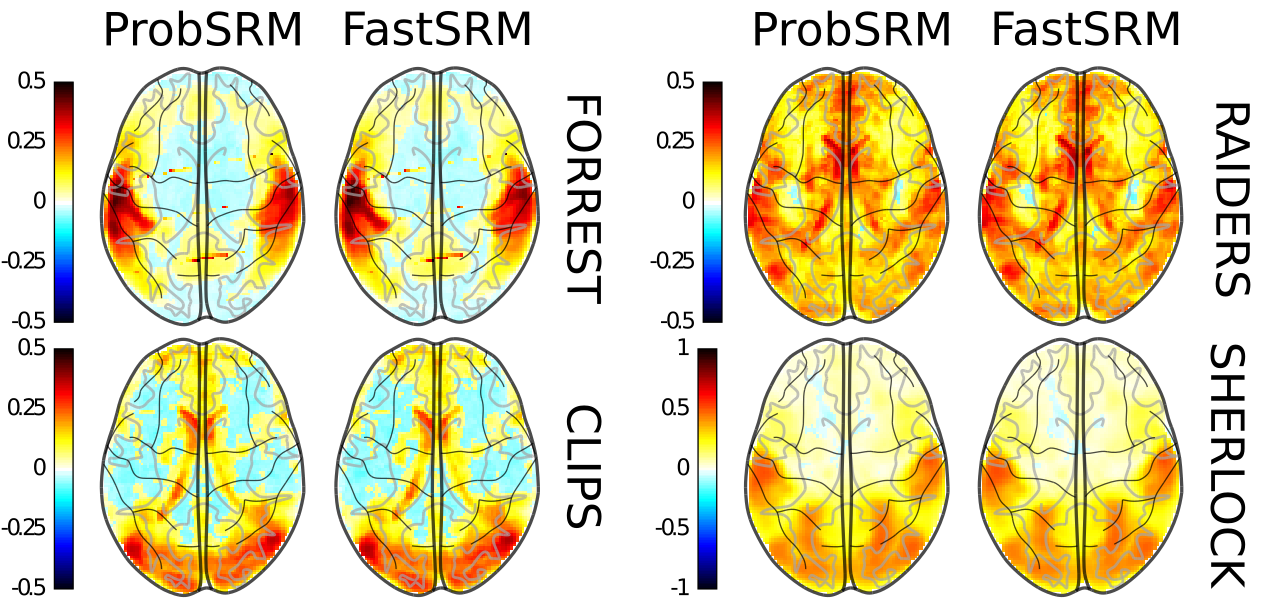}
\caption{\textbf{fMRI reconstruction: $R^2$ score per voxels averaged across cross validation folds}: We benchmark ProbSRM and FastSRM using $20$ components on SHERLOCK, RAIDERS, CLIPS and FORREST datasets. An $R^2$ score of 1 means perfect reconstruction while 0 means that the model only predicts the mean of the voxel time-course.}
\label{fig:example_r2}
\end{figure}

In Table~\ref{tab:typical}, we show typical values for the main dataset parameters. In Table~\ref{tab:dataset_desc}, we describe the main dataset parameters of the real datasets we used. In Figure~\ref{fig:memory_usage} we compare the memory usage of DetSRM, ProbSRM and FastSRM with different atlases as a function of the number of components used while performing the encoding experiment described in Section~\ref{reconstruction}. In Figure~\ref{fig:example_r2}, we show for the same experiment, the $R^2$ score per voxels averaged across cross validation folds.

\begin{table}
\centering
\begin{tabular}{c|c|c}
	Name of variable &  Notation & Typical value \\
	\hline
	Number of voxels & $v$ & [$10^5, 10^6$] \\
	Number of timeframes & $t$ & [$10^2, 10^3$] \\
	Number of runs & $m$ & $10$ \\
	Number of subjects & $n$ & [$10, 10^2$] \\
	Number of components & $k$ & [$10, 10^2$]
\end{tabular}
\caption{\textbf{Typical values for the main dataset parameters.}}
\label{tab:typical}
\end{table}

\begin{table}
	\begin{tabular}{|c|c|c|c|c|}
		\hline
		\textbf{Dataset} & \textbf{Subjects} & \textbf{Runs} & \textbf{Average run} & \textbf{Voxels} \\
	&& & \textbf{length} & \textbf{(per subject)} \\
	&& & \textbf{(in timeframes)} &  \\
	&$n$& $m$ & $t$ &$v$  \\
		\hline
		CLIPS & 10 & 17 & 325 & 212445\\ 
		\hline
		SHERLOCK & 16 & 5 & 395 & 212445 \\ 
		\hline
		RAIDERS & 10 & 9 & 330 & 212445 \\
		\hline 
		FORREST & 19 & 7 & 465 & 212445\\
		\hline
		CamCAN & 647 & 1 & 193 & 212445 \\
		\hline
	\end{tabular}
\caption{\textbf{Datasets description}}
\label{tab:dataset_desc}
\end{table}

%
\section*{Acknowledgment}
CamCAN Data collection and sharing was provided by the Cambridge Centre for Ageing and Neuroscience (CamCAN). CamCAN funding was provided by the UK Biotechnology and Biological Sciences Research Council (grant number BB/H008217/1), together with support from the UK Medical Research Council and University of Cambridge, UK.

This project has received funding from the European Union's Horizon 2020 Research and Innovation Programme under Grant Agreement No. 785907 (HBP SGA2).

\section*{References}
\bibliography{biblio}

\begin{thebibliography}{10}
\expandafter\ifx\csname url\endcsname\relax
  \def\url#1{\texttt{#1}}\fi
\expandafter\ifx\csname urlprefix\endcsname\relax\def\urlprefix{URL }\fi
\expandafter\ifx\csname href\endcsname\relax
  \def\href#1#2{#2} \def\path#1{#1}\fi

\bibitem{poline2012general}
J.-B. Poline, M.~Brett, The general linear model and fmri: does love last
  forever?, Neuroimage 62~(2) (2012) 871--880.

\bibitem{huth2012continuous}
A.~G. Huth, S.~Nishimoto, A.~T. Vu, J.~L. Gallant, A continuous semantic space
  describes the representation of thousands of object and action categories
  across the human brain, Neuron 76~(6) (2012) 1210--1224.

\bibitem{gucclu2017increasingly}
U.~G{\"u}{\c{c}}l{\"u}, M.~A. van Gerven, Increasingly complex representations
  of natural movies across the dorsal stream are shared between subjects,
  NeuroImage 145 (2017) 329--336.

\bibitem{eickenberg2017seeing}
M.~Eickenberg, A.~Gramfort, G.~Varoquaux, B.~Thirion, Seeing it all:
  Convolutional network layers map the function of the human visual system,
  NeuroImage 152 (2017) 184--194.

\bibitem{richard2018optimizing}
H.~Richard, A.~Pinho, B.~Thirion, G.~Charpiat, Optimizing deep video
  representation to match brain activity, arXiv preprint arXiv:1809.02440.

\bibitem{gucclu2015deep}
U.~G{\"u}{\c{c}}l{\"u}, M.~A. van Gerven, Deep neural networks reveal a
  gradient in the complexity of neural representations across the ventral
  stream, Journal of Neuroscience 35~(27) (2015) 10005--10014.

\bibitem{hasson2004intersubject}
U.~Hasson, Y.~Nir, I.~Levy, G.~Fuhrmann, R.~Malach, Intersubject
  synchronization of cortical activity during natural vision, science
  303~(5664) (2004) 1634--1640.

\bibitem{chen2015reduced}
P.-H.~C. Chen, J.~Chen, Y.~Yeshurun, U.~Hasson, J.~Haxby, P.~J. Ramadge, A
  reduced-dimension fmri shared response model, in: Advances in Neural
  Information Processing Systems, 2015, pp. 460--468.

\bibitem{anderson2016enabling}
M.~J. Anderson, M.~Capota, J.~S. Turek, X.~Zhu, T.~L. Willke, Y.~Wang, P.-H.
  Chen, J.~R. Manning, P.~J. Ramadge, K.~A. Norman, Enabling factor analysis on
  thousand-subject neuroimaging datasets, in: 2016 IEEE International
  Conference on Big Data (Big Data), IEEE, 2016, pp. 1151--1160.

\bibitem{vodrahalli2018mapping}
K.~Vodrahalli, P.-H. Chen, Y.~Liang, C.~Baldassano, J.~Chen, E.~Yong, C.~Honey,
  U.~Hasson, P.~Ramadge, K.~A. Norman, et~al., Mapping between fmri responses
  to movies and their natural language annotations, Neuroimage 180 (2018)
  223--231.

\bibitem{chen2017shared}
J.~Chen, Y.~C. Leong, C.~J. Honey, C.~H. Yong, K.~A. Norman, U.~Hasson, Shared
  memories reveal shared structure in neural activity across individuals,
  Nature neuroscience 20~(1) (2017) 115.

\bibitem{mckeown1998analysis}
M.~J. McKeown, S.~Makeig, G.~G. Brown, T.-P. Jung, S.~S. Kindermann, A.~J.
  Bell, T.~J. Sejnowski, Analysis of fmri data by blind separation into
  independent spatial components, Human brain mapping 6~(3) (1998) 160--188.

\bibitem{beckmann2004probabilistic}
C.~F. Beckmann, S.~M. Smith, Probabilistic independent component analysis for
  functional magnetic resonance imaging, IEEE transactions on medical imaging
  23~(2) (2004) 137--152.

\bibitem{calhoun2001method}
V.~D. Calhoun, T.~Adali, G.~D. Pearlson, J.~J. Pekar, A method for making group
  inferences from functional mri data using independent component analysis,
  Human brain mapping 14~(3) (2001) 140--151.

\bibitem{varoquaux2010group}
G.~Varoquaux, S.~Sadaghiani, P.~Pinel, A.~Kleinschmidt, J.-B. Poline,
  B.~Thirion, A group model for stable multi-subject ica on fmri datasets,
  Neuroimage 51~(1) (2010) 288--299.

\bibitem{beckmann2005tensorial}
C.~F. Beckmann, S.~M. Smith, Tensorial extensions of independent component
  analysis for multisubject fmri analysis, Neuroimage 25~(1) (2005) 294--311.

\bibitem{smith2004advances}
S.~M. Smith, M.~Jenkinson, M.~W. Woolrich, C.~F. Beckmann, T.~E. Behrens,
  H.~Johansen-Berg, P.~R. Bannister, M.~De~Luca, I.~Drobnjak, D.~E. Flitney,
  et~al., Advances in functional and structural mr image analysis and
  implementation as fsl, Neuroimage 23 (2004) S208--S219.

\bibitem{xie2017decoding}
J.~Xie, P.~K. Douglas, Y.~N. Wu, A.~L. Brody, A.~E. Anderson, Decoding the
  encoding of functional brain networks: An fmri classification comparison of
  non-negative matrix factorization (nmf), independent component analysis
  (ica), and sparse coding algorithms, Journal of neuroscience methods 282
  (2017) 81--94.

\bibitem{varoquaux:inria-00588898}
G.~Varoquaux, A.~Gramfort, F.~Pedregosa, V.~Michel, B.~Thirion,
  \href{https://hal.inria.fr/inria-00588898}{{Multi-subject dictionary learning
  to segment an atlas of brain spontaneous activity}}, in: {Information
  Processing in Medical Imaging}, Vol. 6801 of Lecture Notes in Computer
  Science, {G{\'a}bor Sz{\'e}kely, Horst Hahn}, {Springer}, Kaufbeuren,
  Germany, 2011, pp. 562--573.
\newblock \href {http://dx.doi.org/10.1007/978-3-642-22092-0\_46}
  {\path{doi:10.1007/978-3-642-22092-0\_46}}.
\newline\urlprefix\url{https://hal.inria.fr/inria-00588898}

\bibitem{abraham}
A.~Abraham, E.~Dohmatob, B.~Thirion, D.~Samaras, G.~Varoquaux, Extracting brain
  regions from rest fmri with total-variation constrained dictionary learning,
  in: International Conference on Medical Image Computing and Computer-Assisted
  Intervention, Springer, 2013, pp. 607--615.

\bibitem{dohmatob2016learning}
E.~Dohmatob, A.~Mensch, G.~Varoquaux, B.~Thirion, Learning brain regions via
  large-scale online structured sparse dictionary learning, in: Advances in
  Neural Information Processing Systems, 2016, pp. 4610--4618.

\bibitem{mensch2016dictionary}
A.~Mensch, J.~Mairal, B.~Thirion, G.~Varoquaux, Dictionary learning for massive
  matrix factorization, in: International Conference on Machine Learning, 2016,
  pp. 1737--1746.

\bibitem{varoquaux2013cohort}
G.~Varoquaux, Y.~Schwartz, P.~Pinel, B.~Thirion, Cohort-level brain mapping:
  learning cognitive atoms to single out specialized regions, in: International
  Conference on Information Processing in Medical Imaging, Springer, 2013, pp.
  438--449.

\bibitem{calhoun2004alcohol}
V.~D. Calhoun, J.~J. Pekar, G.~D. Pearlson, Alcohol intoxication effects on
  simulated driving: exploring alcohol-dose effects on brain activation using
  functional mri, Neuropsychopharmacology 29~(11) (2004) 2097.

\bibitem{calhoun2002different}
V.~D. Calhoun, J.~J. Pekar, V.~B. McGinty, T.~Adali, T.~D. Watson, G.~D.
  Pearlson, Different activation dynamics in multiple neural systems during
  simulated driving, Human brain mapping 16~(3) (2002) 158--167.

\bibitem{lashkari2009exploratory}
D.~Lashkari, P.~Golland, Exploratory fmri analysis without spatial
  normalization, in: International Conference on Information Processing in
  Medical Imaging, Springer, 2009, pp. 398--410.

\bibitem{guntupalli2016model}
J.~S. Guntupalli, M.~Hanke, Y.~O. Halchenko, A.~C. Connolly, P.~J. Ramadge,
  J.~V. Haxby, A model of representational spaces in human cortex, Cerebral
  cortex 26~(6) (2016) 2919--2934.

\bibitem{DBLP:conf/ipmi/BazeilleRJT19}
T.~Bazeille, H.~Richard, H.~Janati, B.~Thirion,
  \href{https://doi.org/10.1007/978-3-030-20351-1\_18}{Local optimal transport
  for functional brain template estimation}, in: Information Processing in
  Medical Imaging - 26th International Conference, {IPMI} 2019, Hong Kong,
  China, June 2-7, 2019, Proceedings, 2019, pp. 237--248.
\newblock \href {http://dx.doi.org/10.1007/978-3-030-20351-1\_18}
  {\path{doi:10.1007/978-3-030-20351-1\_18}}.
\newline\urlprefix\url{https://doi.org/10.1007/978-3-030-20351-1\_18}

\bibitem{chen2016convolutional}
P.-H. Chen, X.~Zhu, H.~Zhang, J.~S. Turek, J.~Chen, T.~L. Willke, U.~Hasson,
  P.~J. Ramadge, A convolutional autoencoder for multi-subject fmri data
  aggregation, arXiv preprint arXiv:1608.04846.

\bibitem{shvartsman2017matrix}
M.~Shvartsman, N.~Sundaram, M.~C. Aoi, A.~Charles, T.~C. Wilke, J.~D. Cohen,
  Matrix-normal models for fmri analysis, arXiv preprint arXiv:1711.03058.

\bibitem{turek2018capturing}
J.~S. Turek, C.~T. Ellis, L.~J. Skalaban, N.~B. Turk-Browne, T.~L. Willke,
  Capturing shared and individual information in fmri data, in: 2018 IEEE
  International Conference on Acoustics, Speech and Signal Processing (ICASSP),
  IEEE, 2018, pp. 826--830.

\bibitem{turek2017semi}
J.~S. Turek, T.~L. Willke, P.-H. Chen, P.~J. Ramadge, A semi-supervised method
  for multi-subject fmri functional alignment, in: 2017 IEEE International
  Conference on Acoustics, Speech and Signal Processing (ICASSP), IEEE, 2017,
  pp. 1098--1102.

\bibitem{bellec2010multi}
P.~Bellec, P.~Rosa-Neto, O.~C. Lyttelton, H.~Benali, A.~C. Evans, Multi-level
  bootstrap analysis of stable clusters in resting-state fmri, Neuroimage
  51~(3) (2010) 1126--1139.

\bibitem{schaefer2017local}
A.~Schaefer, R.~Kong, E.~M. Gordon, T.~O. Laumann, X.-N. Zuo, A.~J. Holmes,
  S.~B. Eickhoff, B.~T. Yeo, Local-global parcellation of the human cerebral
  cortex from intrinsic functional connectivity mri, Cerebral Cortex 28~(9)
  (2017) 3095--3114.

\bibitem{mensch2018extracting}
A.~Mensch, J.~Mairal, B.~Thirion, G.~Varoquaux, Extracting universal
  representations of cognition across brain-imaging studies, arXiv preprint
  arXiv:1809.06035.

\bibitem{abraham2014machine}
A.~Abraham, F.~Pedregosa, M.~Eickenberg, P.~Gervais, A.~Mueller, J.~Kossaifi,
  A.~Gramfort, B.~Thirion, G.~Varoquaux, Machine learning for neuroimaging with
  scikit-learn, Frontiers in neuroinformatics 8 (2014) 14.

\bibitem{sherlock}
J.~Chen, Y.~Leong, K.~Norman, U.~Hasson,
  \href{https://www.biorxiv.org/content/early/2016/01/05/035931}{Shared
  experience, shared memory: a common structure for brain activity during
  naturalistic recall}, bioRxiv\href
  {http://arxiv.org/abs/https://www.biorxiv.org/content/early/2016/01/05/035931.full.pdf}
  {\path{arXiv:https://www.biorxiv.org/content/early/2016/01/05/035931.full.pdf}},
  \href {http://dx.doi.org/10.1101/035931} {\path{doi:10.1101/035931}}.
\newline\urlprefix\url{https://www.biorxiv.org/content/early/2016/01/05/035931}

\bibitem{poldrack2013toward}
R.~A. Poldrack, D.~M. Barch, J.~Mitchell, T.~Wager, A.~D. Wagner, J.~T. Devlin,
  C.~Cumba, O.~Koyejo, M.~Milham, Toward open sharing of task-based fmri data:
  the openfmri project, Frontiers in neuroinformatics 7 (2013) 12.

\bibitem{hanke2014high}
M.~Hanke, F.~J. Baumgartner, P.~Ibe, F.~R. Kaule, S.~Pollmann, O.~Speck,
  W.~Zinke, J.~Stadler, A high-resolution 7-tesla fmri dataset from complex
  natural stimulation with an audio movie, Scientific data 1 (2014) 140003.

\bibitem{ibc}
A.~L. Pinho, A.~Amadon, T.~Ruest, M.~Fabre, E.~Dohmatob, I.~Denghien,
  C.~Ginisty, S.~Becuwe-Desmidt, S.~Roger, L.~Laurier, et~al., Individual brain
  charting, a high-resolution fmri dataset for cognitive mapping, Scientific
  data 5.

\bibitem{haxby2011common}
J.~V. Haxby, J.~S. Guntupalli, A.~C. Connolly, Y.~O. Halchenko, B.~R. Conroy,
  M.~I. Gobbini, M.~Hanke, P.~J. Ramadge, A common, high-dimensional model of
  the representational space in human ventral temporal cortex, Neuron 72~(2)
  (2011) 404--416.

\bibitem{nishimoto2011reconstructing}
S.~Nishimoto, A.~Vu, T.~Naselaris, Y.~Benjamini, B.~Yu, J.~L. Gallant,
  Reconstructing visual experiences from brain activity evoked by natural
  movies, Current Biology 21~(19) (2011) 1641--1646.

\bibitem{taylor2017cambridge}
J.~R. Taylor, N.~Williams, R.~Cusack, T.~Auer, M.~A. Shafto, M.~Dixon, L.~K.
  Tyler, R.~N. Henson, et~al., The cambridge centre for ageing and neuroscience
  (cam-can) data repository: structural and functional mri, meg, and cognitive
  data from a cross-sectional adult lifespan sample, Neuroimage 144 (2017)
  262--269.

\bibitem{shafto2014cambridge}
M.~A. Shafto, L.~K. Tyler, M.~Dixon, J.~R. Taylor, J.~B. Rowe, R.~Cusack, A.~J.
  Calder, W.~D. Marslen-Wilson, J.~Duncan, T.~Dalgleish, et~al., The cambridge
  centre for ageing and neuroscience (cam-can) study protocol: a
  cross-sectional, lifespan, multidisciplinary examination of healthy cognitive
  ageing, BMC neurology 14~(1) (2014) 204.

\bibitem{wu2018learning}
A.~Wu, S.~Pashkovski, S.~R. Datta, J.~W. Pillow, Learning a latent manifold of
  odor representations from neural responses in piriform cortex, in: Advances
  in Neural Information Processing Systems, 2018, pp. 5378--5388.

\bibitem{rahim2017joint}
M.~Rahim, B.~Thirion, D.~Bzdok, I.~Buvat, G.~Varoquaux, Joint prediction of
  multiple scores captures better individual traits from brain images,
  NeuroImage 158 (2017) 145--154.

\bibitem{pedregosa2011scikit}
F.~Pedregosa, G.~Varoquaux, A.~Gramfort, V.~Michel, B.~Thirion, O.~Grisel,
  M.~Blondel, P.~Prettenhofer, R.~Weiss, V.~Dubourg, et~al., Scikit-learn:
  Machine learning in python, Journal of machine learning research 12~(Oct)
  (2011) 2825--2830.

\bibitem{breiman2001random}
L.~Breiman, Random forests, Machine learning 45~(1) (2001) 5--32.

\bibitem{louppe2013understanding}
G.~Louppe, L.~Wehenkel, A.~Sutera, P.~Geurts, Understanding variable
  importances in forests of randomized trees, in: Advances in neural
  information processing systems, 2013, pp. 431--439.

\bibitem{bijsterbosch2018relationship}
J.~D. Bijsterbosch, M.~W. Woolrich, M.~F. Glasser, E.~C. Robinson, C.~F.
  Beckmann, D.~C. Van~Essen, S.~J. Harrison, S.~M. Smith, The relationship
  between spatial configuration and functional connectivity of brain regions,
  Elife 7 (2018) e32992.

\end{thebibliography}
\end{document}